%% file: main.tex
\documentclass[sigconf]{acmart}

\usepackage{algorithm}
\usepackage{algorithmic}
\usepackage{amsthm,amsmath}
\usepackage{enumitem}
\usepackage{multirow}
\usepackage{multicol}
\usepackage{tcolorbox}
\newtheorem{problem}{Problem}
\newtheorem{assumption}{Assumption}
\newtheorem{lemma}{Lemma}

\AtBeginDocument{%
  \providecommand\BibTeX{{%
    \normalfont B\kern-0.5em{\scshape i\kern-0.25em b}\kern-0.8em\TeX}}}

\copyrightyear{2022} 
\acmYear{2022} 
\setcopyright{acmcopyright}\acmConference[WWW '22]{Proceedings of the ACM Web Conference 2022}{April 25--29, 2022}{Virtual Event, Lyon, France}
\acmBooktitle{Proceedings of the ACM Web Conference 2022 (WWW '22), April 25--29, 2022, Virtual Event, Lyon, France}
\acmPrice{15.00}
\acmDOI{10.1145/3485447.3511969}
\acmISBN{978-1-4503-9096-5/22/04}

\makeatletter
\def\maketag@@@#1{\hbox{\m@th\normalfont\normalsize#1}}
\makeatother

\title{CausPref: Causal Preference Learning for Out-of-Distribution Recommendation}

\author[Yue He, Zimu Wang, Peng Cui, Hao Zou, Yafeng Zhang, Qiang Cui and Yong Jiang]{Yue He$^{1*}$, Zimu Wang$^{1*}$, Peng Cui$^{1\dagger}$, Hao Zou$^1$, Yafeng Zhang$^2$, Qiang Cui$^2$, Yong Jiang$^1$}
\affiliation{ 
  \institution{$^1$Department of Computer Science and Technology, Tsinghua University, Beijing, China}
  \country{}
}
\affiliation{ 
  \institution{$^2$Meituan, Beijing, China}
  \country{}
}
\email{heyue18@mails.tsinghua.edu.cn, 14317593@qq.com, cuip@tsinghua.edu.cn}
\email{ahio@163.com, {zhangyafeng, cuiqiang04}@meituan.com, jiangy@sz.tsinghua.edu.cn}

\renewcommand{\authors}{Yue He, Zimu Wang, Peng Cui, Hao Zou, Yafeng Zhang, Qiang Cui and Yong Jiang}

\begin{document}
\begin{abstract}
    \renewcommand{\thefootnote}{\fnsymbol{footnote}}
    \footnotetext[1]{Equal Contributions}
    \footnotetext[2]{Corresponding Author}
  In spite of the tremendous development of recommender system owing to the progressive capability of machine learning recently, the current recommender system is still vulnerable to the distribution shift of users and items in realistic scenarios, leading to the sharp decline of performance in testing environments. It is even more severe in many common applications where only the implicit feedback from sparse data is available. Hence, it is crucial to promote the performance stability of recommendation method in different environments. In this work, we first make a thorough analysis of implicit recommendation problem from the viewpoint of out-of-distribution (OOD) generalization. Then under the guidance of our theoretical analysis, we propose to incorporate the recommendation-specific DAG learner into a novel 
  causal preference-based recommendation framework named CausPref, mainly consisting of causal learning of invariant user preference and anti-preference negative sampling to deal with implicit feedback. Extensive experimental results\footnotetext[3]{https://github.com/HeYueThu/CausPref}  from real-world datasets clearly demonstrate that our approach surpasses the benchmark models significantly under types of out-of-distribution settings, and show its impressive interpretability. 
\end{abstract}




\maketitle

\input{paragraph/introduction}
\input{paragraph/related_work}
\input{paragraph/algorithm}
\input{paragraph/experiment}
\input{paragraph/conclusion}

\section*{Acknowledgments}

This work was supported in part by National Key R\&D Program
of China (No. 2018AAA0102004), National
Natural Science Foundation of China (No. U1936219, 62141607), and Beijing Academy of Artificial Intelligence (BAAI).

\clearpage
\bibliographystyle{ACM-Reference-Format}
\bibliography{main}

\clearpage
\setcounter{table}{0}
\input{paragraph/appendix}

\end{document}

%% file: paragraph/introduction.tex
\section{Introduction}
With the growing information overload on website, recommender system plays a pivotal role to alleviate it in online services such as E-commerce and social media site.
To meet user preference, a large amount of recommendation algorithms have been proposed, including collaborative filtering \cite{DBLP:journals/corr/abs-1708-05024}, content-based recommendation \cite{DBLP:conf/recsys/Musto10}, hybrid recommendation \cite{DBLP:conf/wsdm/SantanaS21} and etc.
During the recent years, the powerful capacity of deep neural networks (DNN) \cite{DBLP:journals/iajit/Al-Jawfi09} and graph convolution networks (GCN) \cite{DBLP:conf/iclr/KipfW17} further facilitate the ability and performance of the recommender system.

\begin{figure}\vspace{3pt}
\includegraphics[width=1.0\linewidth]{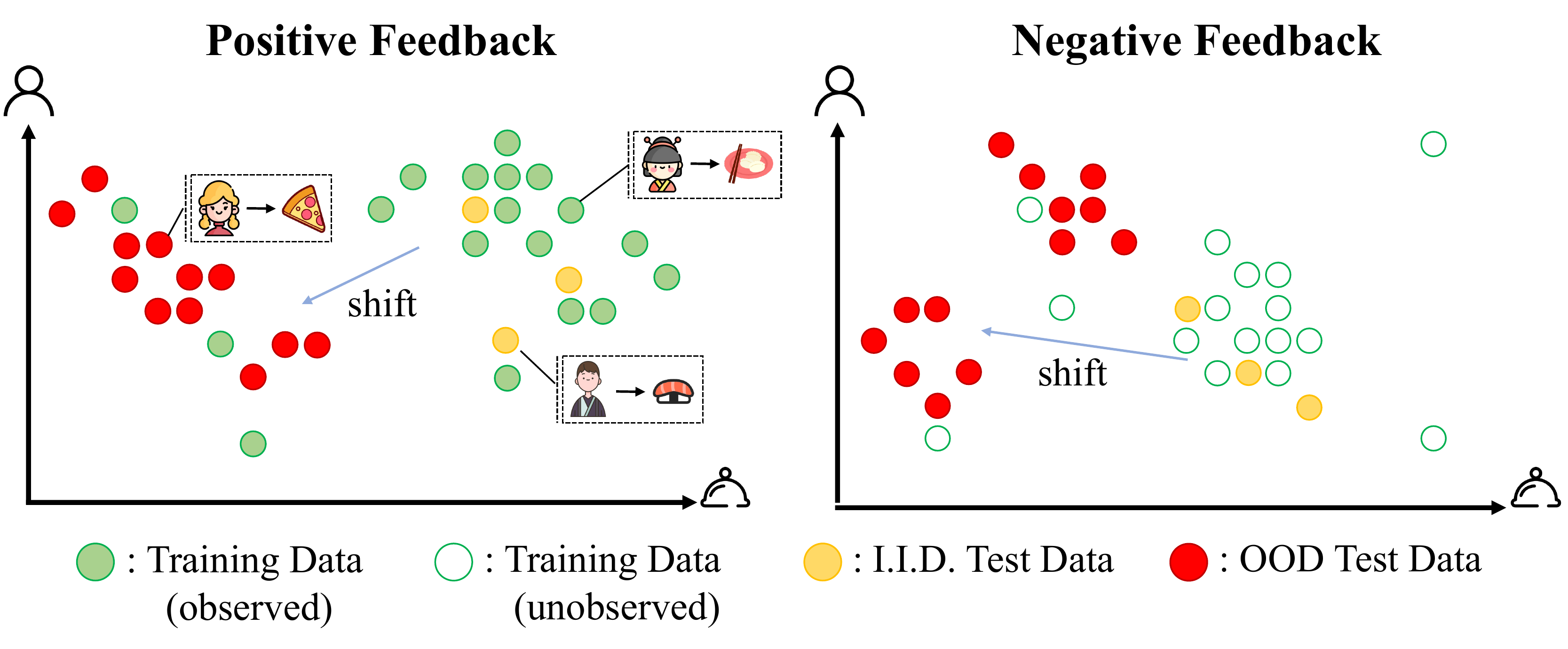}
\centering \vspace{-10pt}
\caption {The Difference of OOD and I.I.D
Recommendation for Implicit Feedback. The distribution shift appears in both the observed positive feedback and the unobserved negative feedback, resulting in the poor performance in OOD test.}
\label{showcase}
\vspace{-10pt}
\end{figure}

Despite their promising performances, most of the current recommendation systems assume independent and identically distributed (I.I.D.) training and test data. Unfortunately, this hypothesis is often violated in realistic recommendation scenarios where distribution shift problem is inevitable. It mainly comes from two sources in general. The first one is the natural shift in real world due to demographic, spatial and temporal heterogeneity in human behaviors. For instance, the distribution of users in a coastal city may be different from the distribution of users in a inland city, and the distribution of items for sale in winter is likely to be dissimilar with the distribution of items in summer, leading to the distribution shift in population level.
The other one is the artificial biases caused by the mechanism of a recommender system itself. For example, the popular items may have more chances to be exposed than the others, resulting in distribution shift if we care more about the recommendations over the subgroup of unpopular items. Although the second kind is recently studied, e.g. the popularity bias problem \cite{abdollahpouri2017controlling}, the first kind is rarely investigated in literature and there lacks of a unified method to deal with these two kinds of distribution shifts, which motivates us to formulate, analyze and model the problem of out-of-distribution (OOD) recommendation in this paper.


Although the OOD problem has become a research foci in recent studies on learning models such as regression \cite{DBLP:journals/corr/abs-1907-02893} and classification \cite{blanchard2021domain},  it is even more challenging in recommendation due to the implicit feedback problem,i.e. only the positive signal (e.g. a user likes an item) is available while negative signal (e.g. a user dislikes an item) can hardly be obtained in most cases. This renders the difficulty to characterize the distribution of user-item pairs with negative feedback which however also induces distribution shifts.
Hence, it is demanded to address the shift of both observed and unobserved user-item distributions under the implicit feedback in OOD recommendation. 

In this paper, we formulate the recommendation problem as estimating $P(Y|\mathbf{U},\mathbf{V})$\footnote{This formulation is derived from the probabilistic recommendation models \cite{schafer2007collaborative}, and it implies that all the information that determines the feedback has been encoded in the representations of user and item, which are the scenarios we address in this paper. } where $\mathbf{U}$ and $\mathbf{V}$ mean the representations of users and items respectively, and $Y$ indicates the feedback of the user over the item. Similar as the branch of studies in OOD generalization \cite{DBLP:journals/corr/abs-2108-13624}, to achieve the generalization ability against distribution shift, we suppose that the true $P(Y|\mathbf{U},\mathbf{V})$ is invariant across environments, while $P(\mathbf{U},\mathbf{V})$ may shift from training to test. We theoretically prove that the invariance of user preferences (i.e. $P(\mathbf{V}|\mathbf{U}, Y=1)$) is a sufficient condition to guarantee the invariance of $P(Y|\mathbf{U},\mathbf{V})$ under reasonable assumptions. Therefore, how to learn invariant user preferences becomes our major target. A large body of literature in invariant learning \cite{buhlmann2020invariance,DBLP:conf/icml/OberstTPS21} have revealed that the causal structure reflecting data generation process can keep invariant under typical data distribution shifts. But how to learn the causal structure of user preference (rather than exploiting a known causal structure to alleviate some specific form of bias as in \cite{zheng2021disentangling}) is still an open problem. 

Inspired by the recent progress in differentiatable causal discovery \cite{DBLP:conf/nips/ZhengARX18}, we propose a new Directed-Acyclic-Graph (DAG) learning method for recommendation, and organically integrate it into a neural collaborative filtering framework, forming a novel causal preference-based recommendation framework
named CausPref. More specifically, the DAG learner is regularized according to the nature of user preference, e.g. only user features may cause item features, but not vice versa. 
Then we can employ it to learn the causal structure of user preference.
To deal with the implicit feedback problem, we also design an anti-preference negative sampling, which makes the framework consistent with our theoretical analysis. Note that we incorporate user features and item features into the our framework to accommodate inductive setting\footnote{During training, we are aware of the users and items for test in transductive learning, and yet new users/items for test exist in inductive learning} because the capability to tackle new users and new items are much more demanded in OOD recommendation than in traditional setting. Besides, the inherent explainability of causality endows CausPref with a side advantage of providing explanations for recommendations.

The main contributions of our paper are as follows: 

\begin{itemize}[leftmargin=0.3cm]
    \item We theoretically analyze the recommendation problem under implicit feedback in OOD generalization perspective.
       \item We propose a novel causal preference-based recommendation framework (CausPref), mainly consisting of causal learning of
       invariant user preference and anti-preference negative sampling to deal with implicit feedback.
    \item We conduct extensive experiments and demonstrate the effectiveness and advantages of our approach in tackling a variety of OOD scenarios.
\end{itemize}

%% file: paragraph/related_work.tex
\section{Related Works}
\subsection{Recommender System} 
The content-based recommendation (CB) and collaborative filtering (CF) are two major approaches in the early study of recommender system. 
Compared to the CB methods \cite{DBLP:conf/recsys/Musto10} that ignore the users' dependence, the CF becomes the mainstream method eventually since the popularity of matrix factorization (MF) \cite{DBLP:journals/corr/abs-1708-05024}, because it can utilize the user-item interactions based on the nearest neighbor. 
Later on, several works (e.g. SVD++ \cite{koren2008factorization}, Factorization Machine \cite{DBLP:conf/www/SunPZF21}) propose to enhance the latent representations of user and item.
With the rise of deep learning \cite{DBLP:journals/iajit/Al-Jawfi09}, the neural network based CF (NCF \cite{DBLP:conf/www/HeLZNHC17}) 
leverages the multi-layer perceptron (MLP) \cite{gardner1998artificial} to model user behaviors. Further, the NeuMF \cite{DBLP:conf/www/HeLZNHC17} extends NCF and combines it with the linear operation of MF. With the user-item graph data, the LightGCN \cite{DBLP:conf/sigir/0001DWLZ020} simplifies the design of GCN \cite{DBLP:conf/iclr/KipfW17} to make it more appropriate for recommendation. To augment the input, the knowledge-based methods \cite{DBLP:conf/kdd/ZhouZBZWY20} use the costly external information, such as knowledge graph, as important supplementary.
Then the hybrid recommendation \cite{DBLP:conf/wsdm/SantanaS21} aggregates multiple types of recommender systems together.

To deal with the data bias problems \cite{chen2020bias} in recommender system, 
a range of research effects have been made.
For example, the IPS-based methods \cite{schnabel2016recommendations, yang2018unbiased} employ the estimated inverse propensity score from biased and unbiased groups to reweight imbalanced samples;
some causal modeling methods \cite{bonner2018causal, zheng2021disentangling} build their models relying on predefined causal structure from domain knowledge of experts; and other approaches utilize and develop the distinct de-bias techniques including \cite{pedreshi2008discrimination, bressan2016limits, abdollahpouri2017controlling, bose2019compositional, wen2020entire} and etc.
However, they lack the general formulation for distribution shift, 
hence can only address the bias of specific form in transductive setting.
Moreover, the requirement on prior knowledge of test environment limits the transfer learning-based \cite{DBLP:conf/cikm/WuCCDRKAZASCCLK20} and RL/bandit-based \cite{DBLP:conf/sigir/XieYZL21} methods to deal with the  agnostic distribution shift.
To pursuit the stable performance for agnostic distribution, in this paper we propose to capture the essential invariant mechanism from user behavior data to solve the common distribution shift problems, and design a model based on available features (referring to the literature of cold start \cite{gantner2010learning}) because the OOD recommendation usually accompanies the emergency of new users/items. Along with that, the causal knowledge learning module in our model contributes to the interpretable recommender system \cite{wang2018tem} as well.

\subsection{Causal Structure Learning}
The aim of causal discovery is to explore the DAG structure that underlies the generation mechanism of variables. 
But the chief obstacle of implementing traditional approaches in real applications is the daunting cost of searching in combinatorial solution space \cite{spirtes2000causation,chickering2002optimal,spirtes2013causal}.
To overcome this limitation, the recent methods solve the problem in continuous program using differentiable optimization with gradient-based acyclicity constraint \cite{DBLP:conf/nips/ZhengARX18}, and further pay attention to the complicated structural equation \cite{DBLP:conf/aistats/ZhengDARX20}, the convergence efficiency \cite{DBLP:conf/nips/NgG020}, the optimization technique \cite{DBLP:conf/iclr/ZhuNC20} and etc. 
In this way, they provide the opportunity to integrate causal discovery into common machine learning tasks. Then CASTLE \cite{DBLP:conf/nips/KyonoZS20} first learns the DAG structure of all the input variables and takes the direct parents of target variable as the predictor in regression or classification task.
As a result, the expected reconstruction loss of variables can be theoretically upper bounded by the empirically loss of finite observational samples.
Compared to CASTLE, this work introduces and elaborately adapts the DAG constraint to meet the specific graph characteristics in recommendation scenario to facilitate both OOD generalization and interpretability.

\subsection{Out-Of-Distribution Generalization}
In recent years, the OOD problem has received a growing attention because it can advance the trustworthy artificial intelligence \cite{thiebes2021trustworthy}. 
A variety of research areas, including computer vision \cite{blanchard2021domain}, natural language processing \cite{borkan2019nuanced} and etc \cite{hu2020open}, have considered the OOD generalization to devise stable models. 
However, to the best of our knowledge, there is no work studying the general OOD problem in recommender system, although the distribution shift of users and items is almost inevitable in realistic scenarios, especially for the sparse implicit feedback data.

%% file: paragraph/algorithm.tex
\section{Problem Statement and Approach}
In this section, we first introduce the problem formulation, then give a detailed analysis of OOD generalization, and a feasible solution called causal preference-based recommendation (CausPref).

\subsection{Out-Of-Distribution Recommendation}
In this work, we study the recommendation problem where distribution shift exists. 
For the most common implicit feedback, the user-item interaction can be represented as $(\mathbf{u}_i, \mathbf{v}_i, y_i)$, where $\mathbf{u}_i \in \mathbb{R}^{1 \times q^{u}}$ is a $q^{u}$ dimensional vector denoting the user by his/her latent feature, $\mathbf{v}_i \in \mathbb{R}^{1 \times q^{v}}$ is a $q^{v}$ dimensional vector denoting the item by its latent feature, and $y_i=1/0$ denotes the feedback presenting the user's preference on the item $\mathbf{v}_i$. Then we can define the out-of-distribution implicit recommendation problem as follow.

\noindent\rule[+0pt]{\linewidth}{0.1em}
\begin{problem}\label{problem1} 
\vspace{-5pt}
Given \ \ the \ \ observational \ \ training \ \ data \ \ $D_{train}=$ \\$\{(\mathbf{u}_i, \mathbf{v}_i, 1)|(\mathbf{u}_i, \mathbf{v}_i)\sim P(\mathbf{U}, \mathbf{V}|Y=1)\}_{i=1}^{N}$
and test data $D_{test}=\{(\mathbf{u}_i^*, \mathbf{v}_i^*, y_i^*)| (\mathbf{u}_i^*, \mathbf{v}_i^*, y_i^*) \sim P^*(\mathbf{U}, \mathbf{V}, Y)\}_{i=1}^{N^*}$, 
where $P(\mathbf{U}, \mathbf{V}) \neq P^*(\mathbf{U}, \mathbf{V})$,
the support of $P^*(\mathbf{U}, \mathbf{V})$ is contained in the support of $P(\mathbf{U}, \mathbf{V})$\footnote{It means that for $\forall \mathbf{u} \in \mathbb{R}^{1 \times q^{u}}
\mathbf{v} \in \mathbb{R}^{1 \times q^{v}}$ that $P^*(\mathbf{U}=\mathbf{u}, \mathbf{V}=\mathbf{v}) > 0$,  we have $P(\mathbf{U}=\mathbf{u}, \mathbf{V}=\mathbf{v}) > 0$, ensuring the feasibility of generalization from training distribution to test distribution.}, 
and $N/N^*$ is the sample sizes of training/test data.
The goal of out-of-distribution implicit recommendation is to learn a recommender system from training distribution $P(\mathbf{U}, \mathbf{V})$ to predict the feedback of the user on the item in test distribution $P^*(\mathbf{U}, \mathbf{V})$ precisely.
\vspace{-2pt}
\end{problem}
\noindent\rule[+3pt]{\linewidth}{0.1em}

According to Problem\ref{problem1}, we can only obtain the partial information of distribution $P(\mathbf{U}, \mathbf{V}, Y)$, i.e. positive feedback  $P(\mathbf{U}, \mathbf{V}, Y=1)$. And to achieve the generalization ability in $P^*(\mathbf{U}, \mathbf{V})$, we suppose to make the Assumption\ref{assumption1} to support the OOD recommendation.

\begin{assumption}\label{assumption1}
The generation mechanism of user-item interaction $P(Y|\mathbf{U}, \mathbf{V})$ is invariant to different environments, that is $P(Y|\mathbf{U}, \mathbf{V})$ always equals to $P^*(Y|\mathbf{U}, \mathbf{V})$, even though $P(\mathbf{U}, \mathbf{V}) \neq P^*(\mathbf{U}, \mathbf{V})$.
\end{assumption}

\subsection{Theoretical Reflection of Generalization}
The objective of recommendation method is to estimate $P(Y|\mathbf{U}, \mathbf{V})$ accurately. Because $Y$ is binary, we can write $P(Y=1|\mathbf{U}, \mathbf{V})$ as r.h.s in Equation\ref{eq1} according to Bayes theorem.
\vspace{-1pt}
\begin{equation}
    \label{eq1}
     P(Y=1|\mathbf{U}, \mathbf{V}) = \frac{P(\mathbf{U}, \mathbf{V}|Y=1)\cdot P(Y=1)}{\resizebox{0.75\linewidth}{!}{$P(\mathbf{U}, \mathbf{V}|Y=1)\cdot P(Y=1) +P(\mathbf{U}, \mathbf{V}|Y=0)\cdot P(Y=0)$}}
\end{equation}

In recommendation scenario, because the data is generated from the unidirectional selection process (i.e. users select items), then we can rewrite the Equation\ref{eq1} as,
\vspace{-1pt}
\begin{equation}\label{eq2}
\begin{aligned}
    P(Y=1|\mathbf{U}, \mathbf{V}) =        \frac{P(\mathbf{V}|\mathbf{U},Y=1)}{\resizebox{0.75\linewidth}{!}{$ P(\mathbf{V}|\mathbf{U},Y=1)+\frac {P(\mathbf{V}|\mathbf{U},Y=0)\cdot P(\mathbf{U}|Y=0)\cdot P(Y=0)}{P(\mathbf{U}|Y=1)\cdot P(Y=1)}$}}
\end{aligned}
\end{equation}

In implicit feedback, we can only observe the positive feedback, hence 
it is necessary to resort to the negative sampling method to approximate 
the unobserved term $P(\mathbf{V}|\mathbf{U},Y=0)\cdot P(\mathbf{U}|Y=0)\cdot P(Y=0)$. Usually, it fixes an user $\mathbf{u}_j \in \{\mathbf{u}_i\}_{i=1}^N$ first, and then selects an item $\mathbf{v}_k$ $\in \{\mathbf{v}_i\}_{i=1}^N$ based on $\mathbf{u}_j$ according to a sampling strategy. Practically, we empirically uses samples from $P(\mathbf{U}|Y=1)$ to approximate $P(\mathbf{U}|Y=0)$ and simulates $P(\mathbf{V}|\mathbf{U},Y=0)$ by the strategy. Since $\mathcal{C}=\frac{P(Y=0)}{P(Y=1)}$ is a constant that is independent of $P(\mathbf{U}, \mathbf{V})$, we can further write Equation\ref{eq2} as follow, 
\vspace{-1pt}
\begin{equation}
\begin{aligned}
\label{eq3}
    P(Y=1|\mathbf{U}, \mathbf{V}) &= \frac{P(\mathbf{V}|\mathbf{U},Y=1)}{ P(\mathbf{V}|\mathbf{U},Y=1)+\mathcal{C}\cdot P(\mathbf{V}|\mathbf{U},Y=0)}
\end{aligned}
\end{equation}

Because we are agnostic to $P(\mathbf{V}|\mathbf{U},Y=0)$, it is needed to suggest the Assumption\ref{assumption2} on the relationship between $\mathbf{U}$, $\mathbf{V}$ and $Y$.

\begin{assumption}\label{assumption2} The  $P(\mathbf{V}|\mathbf{U},Y=1)$ implied in the observational data is invariant as the $P(\mathbf{U}, \mathbf{V})$ changes, and it can determine the $P(\mathbf{V}|\mathbf{U},Y=0)$ in unobserved  $P(\mathbf{U}, \mathbf{V},Y=0)$, formally $P(\mathbf{V}|\mathbf{U},Y=0)=\mathcal{F}(P(\mathbf{V}|\mathbf{U},Y=1))$.
\end{assumption}

In practice, Assumption\ref{assumption2} is widely accepted in recommendation scenario. $P(\mathbf{V}|\mathbf{U},Y=1)$ and $P(\mathbf{V}|\mathbf{U},Y=0)$ describe the user preference for items, i.e. the item distribution that user likes and dislikes, and the densities of two distributions are usually opposite to each other. 
If all the features concerning recommendation are obtained, the true user preference is supposed to be always invariant. As a result, we can rewrite Equation\ref{eq3} as,
\vspace{-1pt}
\begin{equation}\label{eq4}
\begin{aligned}
    P(Y=1|\mathbf{U}, \mathbf{V}) &= \frac{P(\mathbf{V}|\mathbf{U},Y=1)}{ P(\mathbf{V}|\mathbf{U},Y=1)+\mathcal{C}\cdot \mathcal{F}(P(\mathbf{V}|\mathbf{U},Y=1))}
\end{aligned}
\end{equation}

\begin{figure*} [t]\vspace{0pt}
\centering 

\includegraphics[width=0.85\linewidth]{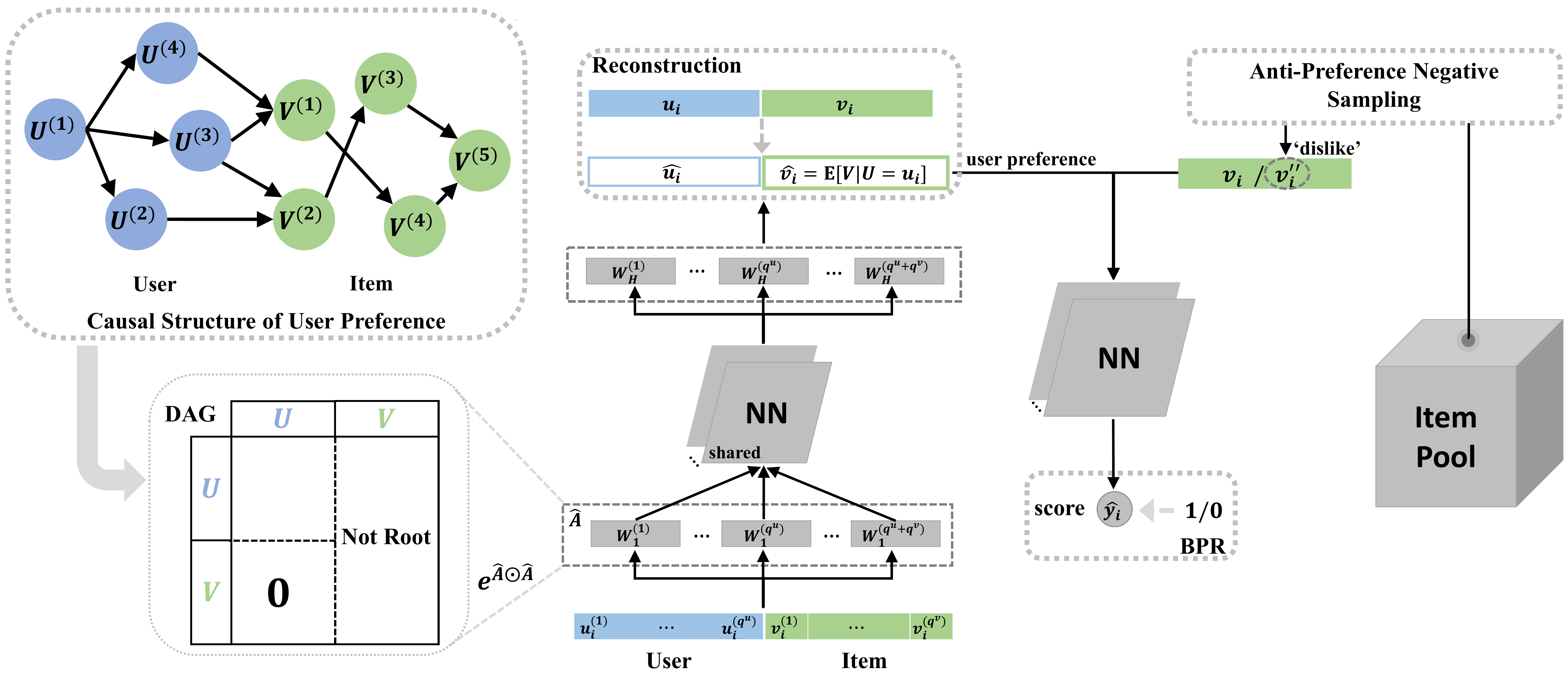}
\centering
\vspace{-5pt}
\caption {The Framework of Causal Preference-based Recommendation.}
\label{fg2}\vspace{-10pt}
\end{figure*}

It can be observed from Equation\ref{eq4} that the target  $P(Y=1|\mathbf{U}, \mathbf{V})$ is only related to $P(\mathbf{V}|\mathbf{U},Y=1)$. This hence guides us a paradigm to achieve the stability of recommender system in two steps:

\begin{tcolorbox}[left=0.0mm]
\begin{itemize} [leftmargin=0.5cm]
    \item Estimate the invariant user preference $P(\mathbf{V}|\mathbf{U},Y=1)$ properly from the observational data $\{(\mathbf{u}_i, \mathbf{v}_i, 1)\}_{i=1}^{N}$.
    \item Estimate the ultimate objective $P(Y=1|\mathbf{U},\mathbf{V})$ from the positive samples, and negative samples selected by a designed strategy $\mathcal{F}(P(\mathbf{V}|\mathbf{U},Y=1))$.
\end{itemize}
\end{tcolorbox}

However, this paradigm is neglected in the framework of traditional recommendation models.
They do not attach importance to the invariant mechanism of user preference $P(\mathbf{V}|\mathbf{U},Y=1)$ well and their negative sampling strategy is independent of $P(\mathbf{V}|\mathbf{U},Y=1)$, e.g. randomly sampling, causing their failed performance in test environment with distribution shift. 
In this paper, we devise a novel causal preference-based recommendation framework named CausPref, which considers the two keypoints together elaborately.
Our CausPref models the $P(\mathbf{V}|\mathbf{U},Y=1)$ explicitly based on the available features and learns it in a causally-regularized way. 

\subsection{Causal Preference Learning}

In this part, we first explain how to utilize the differentiable causal discovery to help to learn the invariant user preference $P(\mathbf{V}|\mathbf{U},Y=1)$ from the observational user-item interactions.

First, we will review the technical details of differentiable causal structure learning.
Let $\mathbf{X}=[\mathbf{X}_1, ..., \mathbf{X}_D]$ denote the $N \times D$ input data matrix sampled from joint distribution $P(X)=P(X_1,...,X_D)$ that is generated from a casual DAG $A$ with $D$ nodes $\{X_1,...,X_D\}$. Then the differentiable approaches for learning a matrix $\hat{A}$  to  approximate $A$ propose to minimize a generic objective function as follow:

\vspace{-10pt}
\begin{equation}\label{eq5}
\min_{\Theta} \mathcal{L}_{dag}=\mathcal{L}_\mathrm{rec}(\hat{A}, \mathbf{X}, \Theta)+\alpha\mathcal{L}_\mathrm{acyc}(\hat{A})+\beta\mathcal{L}_\mathrm{spa}(\hat{A})
\end{equation}

\noindent where $\Theta$ denotes the parameter set of a learnable model $g$, $\mathcal{L}_\mathrm{rec}$ refers to the ability of $g$ to recover the observational data $\mathbf{X}$ based on $\hat{A}$, $\mathcal{L}_\mathrm{acyc}$ is a continuously optimizable acyclicity constraint, and minimizing $\mathcal{L}_\mathrm{spa}$ guarantees the sparsity of $\hat{A}$.

To realize the compliance with data of the causal DAG model, we design a data reconstruction module. Specifically, we set a series of network $g_{\Theta}=\{g^{(d)}\}_{d=1}^D$, each of which reconstruct one element in $\mathbf{X}$. The detailed architecture is shown is Fig.\ref{fg2}. As shown in Fig.\ref{fg2}, we let $\mathbf{W}_1=\{\mathbf{W}_1^{(d)}\in \mathbb{R}^{D\times M}\}_{d=1}^D$ denote the weight matrix set of input layers, $\{\mathbf{W}_h \in \mathbb{R}^{M\times M}\}_{h=2}^{H-1}$  denote the shared weight matrices of middle hidden layers of all sub-networks, and $\mathbf{W}_H=\{\mathbf{W}_H^{(d)}\in \mathbb{R}^{M\times 1}\}_{d=1}^D$ denote the weight matrix set of output layers.
For each $g^{(d)}$ designed to predict the variable $X_d$ (formalized in Equation\ref{aaa}), we enforce the $d$-th row of $\mathbf{W}_1^{(d)}$ to be zero, such that the raw feature $\mathbf{X}_d$ is discarded (not predicting $X_d$ by itself). 
\vspace{-1pt}
\begin{equation}
\label{aaa}
    g^{(d)} = \mathbf{ReLU} (\cdots (\mathbf{ReLU} (\mathbf{X}\cdot \mathbf{W}_1^{(d)}) \cdot \mathbf{W}_2) \cdots \mathbf{W}_{H-1}) \cdot \mathbf{W}_H^{(d)}.
\end{equation}
The reconstruction loss is formulated as
$\mathcal{L}_\mathrm{rec}^N=\frac{1}{N}\|\mathbf{X} - g_{\Theta}(\mathbf{X})\|^2_F$.

To guarantee the matrix $\hat{A}$ of the model, in which $\hat{A}_{k,d}$ is the l2-norm of k-th row of $\mathbf{W}_1^{(d)}$, is a DAG, it is needed to constrain $\mathrm{tr}(\mathrm{e}^{\hat{A}\circ \hat{A}})-d=0$\cite{DBLP:conf/nips/ZhengARX18}. Therefore, we set $\mathcal{L}_{aya} = (\mathrm{tr}(\mathrm{e}^{\hat{A}\circ \hat{A}})-d)^2$ and attempt to minimize it.

To achieve the sparsity of matrix $\hat{A}$, we propose to minimize the L1-norm of matrices $\mathbf{W}_1$.

In summary, we can derive the total loss $\mathcal{L}_{dag}$ term as, 
\vspace{-1pt}
\begin{equation}\label{eq7}
    \begin{aligned}
     \mathcal{L}_{dag}=&\frac{1}{N}\|\mathbf{X} - g_{\Theta}(\mathbf{X})\|^2_F+\alpha(\mathrm{tr}(\mathrm{e}^{\hat{A}\circ \hat{A}})-d)^2 \\& + \beta \|\mathbf{W}_1\|+\chi \|\{\mathbf{W}_h\}_{h=2}^{H}\|^2_2
    \end{aligned}
\end{equation}

\noindent where $\|\cdot\|_F$ is the Febenius norm.

\begin{lemma}\label{le1}
(CASTLE) If  for any sample $\hat{\mathbf{X}} \sim P(X)$, $\|\hat{\mathbf{X}}\|_2 \leq B$ for some $B>0$;
and $\forall \ t>0$, 
$\mathbb{E}_{P(X)}[exp(t(\mathcal{L}_\mathrm{rec} - \mathcal{L}_\mathrm{rec}^P))] \leq exp(\frac{t^2s^2}{2})$, $\mathcal{L}_\mathrm{rec}=\|\mathbf{\hat{\mathbf{X}}} - g_{\Theta}(\mathbf{\hat{\mathbf{X}}})\|^2$;
and each of $\Theta$ has the spectral norm bounded by $\kappa$, then $\forall \delta, \gamma >0$, with probability $1-\delta$, over a training set of $N$ i.i.d. samples, for any $\Theta$, we have:
\vspace{-10pt}
$$\mathcal{L}_\mathrm{rec}^P \le 4 \mathcal{L}_\mathrm{rec}^N + \frac{1}{N}[\mathcal{L}_{aya}+C_1(\mathcal{L}_{spa}+\mathcal{L}_{penalty})+\log(\frac{8}{\delta})]+C_2$$

\noindent where $\mathcal{L}_\mathrm{rec}^P$ is the expected reconstruction loss of $\hat{\mathbf{X}}$\ $\sim$\ $ P(X)$, $\mathcal{L}_\mathrm{rec}^N$ is the empirically  loss on the $N$ training samples
, $C_1$ and $C_2$ are constants depending on $B, s, \gamma, D, H, M$.
\end{lemma}

Lemma\ref{le1} points out that a lower empirical loss of Equation\ref{eq7} means that the learned model is closer to the true causal DAG. In other word, if there exists an underlying causal DAG describing the data generation process, the DAG regularizer can help the model making prediction based on the invariant mechanism \cite{buhlmann2020invariance}, characterized by the causal DAG.  

Inspired by Lemma\ref{le1}, we can leverage the DAG regularizer
to help to learn the invariant user preference $P(\mathbf{V}|\mathbf{U},Y=1)$ with its causal structure from the sparse observational samples $\{(\mathbf{u}_i, \mathbf{v}_i, 1)\}_{i=1}^{N}$. 

First, we define a causal structure of user preference on the feature level of user and item, describing the type of items that user likes.
Differing from the vanilla causal graph, this recommendation-specific causal graph has two additional properties: 

\begin{tcolorbox} [left=0.0mm]
    \begin{itemize}[leftmargin=0.5cm]
    \item C1: Only paths from user features to item features exist. 
    \item C2: Any item feature is not a root node.
\end{itemize}
\end{tcolorbox}

\begin{algorithm}[tb]\label{al1}
\caption{Anti-Preference Negative Sampling (APS)}
\label{alg}
\begin{algorithmic}
    \STATE {\bfseries Input:} Parameter set $\Theta$ to estimate $P(\mathbf{V}|\mathbf{U},Y=1)$, an user $\mathbf{u}_i$ and all items $\{\mathbf{v}_i\}_{i=1}^N$.
    \STATE {\bfseries Step1:} Compute $\mathbb{E} (\mathbf{V}|\mathbf{U}=\mathbf{u}_i,Y=1)$, given $\Theta$.
    \STATE {\bfseries Step2:} Randomly select $K$ items from $\{\mathbf{v}_i\}_{i=1}^N$, then have candidate item set $\{\mathbf{v}_i^{'}\}_{i=1}^K$.
    \STATE {\bfseries Step3:} Compute the sampling weight $\{w_i=\tau_2 (\tau_1(\mathbb{E} (\mathbf{V}|\mathbf{U}=\mathbf{u}_i,Y=1), \mathbf{v}_i^{'}))\}_{i=1}^K$ based on the similarity function $\tau_1$ and reverse operator $\tau_2$.
    \STATE {\bfseries Step4:} Sample an item $\mathbf{v}^{''}_i$ according to the probability $\{\frac{w_i}{\sum_{i=1}^K w_i}\}_{k=1}^K$ of all candidate items.
    \STATE {\bfseries return:} Negative sample $(\mathbf{u}_i, \mathbf{v}^{''}_i)$.
\end{algorithmic}
\end{algorithm}

The two points is accord with people's intuition and the characteristics of recommendation scenario, that is the unidirectional item selection from users implies the preference of active user for the passive item is spontaneous (C1) and depends on the user himself/herself (C2), if all the relevant features are observed. 
Further, with access to the semantics of each features, we can also directly utilize the graph to explain the recommendation process.

To learn the invariant user preference with the DAG regularizer, we concentrate all the features in matrix $\hat{A}$ ($[1:q_u]$ refers to the user and $[q_u+1:q_u+q_v]$ refers to the item) and make the causal structure learning of them.  
To adapt the recommendation-specific properties to promote the learning process,
we add two new term into $\mathcal{L}_{dag}$ as follow,

\begin{equation}\label{eq8}
    \begin{aligned}
    \mathcal{L}_{dag}^{rs}=\mathcal{L}_{dag}&+\xi \mathcal{L}_{zero}(=\|\hat{A}_{[q_u+1:q_u+q_v, 1:q_u]}\|) \\&+ \lambda \mathcal{L}_{nor}(=-\sum_{d=q_u+1}^{q_u+q_v}\log\|\hat{A}_{[:, d]}\|)
    \end{aligned}
\end{equation}

\noindent where $\xi$ is a large constant to ensure $\mathcal{L}_{dag}=0$ that means no path from item to user, and $\mathcal{L}_{nor}$ enforces the no root constraint without hurting the global sparsity, i.e. the number of edges, because it only require the non-zero in-degree of all the item features.

After training, along with the directed path in DAG, 
if one inputs an user's feature $\mathbf{u}_i$, 
the output $\hat{g}_{\Theta}([\mathbf{u}_i])=g_{\Theta}([\mathbf{u}_i|\mathbf{0}])_{[:, q_u+1, q_u+q_v]}$ denotes the expectation of user $\mathbf{u}_i$'s preference actually, that is $\hat{g}_{\Theta}([\mathbf{u}_i])=\mathbb{E} (\mathbf{V}|\mathbf{U}=\mathbf{u}_i,Y=1)$.

\subsection{Causal Preference-based Recommendation}

In this part, we will introduce how to use the estimation of invariant preference $P(\mathbf{V}|\mathbf{U},Y=1)$ to learn the prediction score $P(Y=1|\mathbf{U},\mathbf{V})$ for stable recommendation in proposed framework.

According to the paradigm mentioned above, the first work is to design a negative sampling strategy based on the $P(\mathbf{V}|\mathbf{U},Y=1)$.
Here, we propose a simple but effective strategy as shown in Algorithm\ref{alg}.
After selecting an user with feature $\mathbf{u}_i$, we first compute his/her expectation preference $\mathbb{E} (\mathbf{V}|\mathbf{U}=\mathbf{u}_i,Y=1)$ and then calculate the similarity (the distance function $\tau_1$, e.g. Euclidean distance) between it and other items to select the one with smallest similarity $\mathbf{v}_j$. Considering the computational efficiency problem, we randomly select K items from the whole item pool as candidate set and conduct the item selection above on them. The pair of user $\mathbf{u}_i$ and item $\mathbf{v}_j$ forms the negative sample.

When the positive sample $( \mathbf{u}_i, \mathbf{v}_i)$ from observational data and the negative sample $( \mathbf{u}_i, \mathbf{v}_j)$ selected by APS (Algorithm\ref{alg}) are provided,      
we can then utilize them to train a model $\mu$ (with the parameter set $\Phi$), which outputs the prediction score $P(Y=1|\mathbf{U}, \mathbf{V})$ for ranking.  
Here we choose the neural collaborative filtering framework \cite{DBLP:conf/www/HeLZNHC17} as the predictor and adopt the commonly used Bayesian Personalized Ranking \cite{DBLP:conf/uai/RendleFGS09} (BPR) loss in Equation\ref{eq9} for optimization. 
The BPR loss takes the pair of positive and negative samples as input and encourages the output score of an observed entry to be higher than its unobserved counterparts. 

\vspace{-10pt}
\begin{equation}\label{eq9}
    \min\ \mathcal{L}_0 = - \sum_{i=1}^N \log \sigma (\mu_{\Phi} ( \mathbf{u}_i, \mathbf{v}_i) - \mu_{\Phi} ( \mathbf{u}_i, \mathbf{v}_i^{''}))               
\end{equation} 
\noindent where $\sigma$ is the Sigmoid activation function.

To bring the benefit of learned invariant user preference $P(\mathbf{V}|\mathbf{U},Y=1)$ to the score predictor,
we can directly replace the original user feature with the expectation of learned user preference $\mathbb{E} (\mathbf{V}|\mathbf{U}=\mathbf{u}_i,Y=1)$ in the input of the predictor $\mu$. 
Finally, we propose to integrate the invariant preference estimation and prediction score learning into an unified causal preference-based recommendation framework.

As shown in Figure\ref{fg2}, given a positive sample $(\mathbf{u}_i, \mathbf{v}_i)$, 
after computing $\mathbb{E} (\mathbf{V}|\mathbf{U}=\mathbf{u}_i,Y=1)$ by $g_{\theta}$, the proposed   
CausPref selects $\mathbf{v}_j$ by APS from item pool as negative sample. 
Through stacked layers of neural network $\mu$, we obtain the prediction score of the two samples. For the positive sample $(\mathbf{u}_i, \mathbf{v}_i)$, we minimize the DAG learning loss $\mathcal{L}_{dag}^{rs}$ and the BPR loss simultaneously. For the negative sample $(\mathbf{u}_i, \mathbf{v}_j)$, we only minimize the BPR loss.
Note that we backpropagate the gradient of BPR loss to optimize the parameter $\Theta$ of DAG learner as well, because
it can offer more supervised signals to find the true causal structure in recommendation scenario and help to improve training.
As a result, we jointly optimize $\Theta$ and $\Phi$ to minimize the objective function in Equation \ref{eq10}.

\vspace{-5pt}
\begin{equation}\label{eq10}
    \begin{aligned}
        \min\ \mathcal{L} = &- \sum_{i=1}^N \log \sigma (\mu_{\Phi} ( \hat{g}_{\Theta}([\mathbf{u}_i]), \mathbf{v}_i) - \mu_{\Phi} ( \hat{g}_{\Theta}([\mathbf{u}_i]), \mathbf{v}_i^{''})) \\&+ \zeta \mathcal{L}_{dag}^{rs}   
    \end{aligned}
\end{equation}

%% file: paragraph/experiment.tex
\section{Experiments}

In this section, we compare the effectiveness of our proposed CausPref with benchmark methods in terms of OOD generalization in real-world datasets, and show its impressive interpretability. 

\begin{table*}[tbp]
  \centering
  \caption{The Experimental Results of Degree Bias Settings in the Cloud Theme Click Dataset. }
  \vspace{-5pt}
  \resizebox{0.85\linewidth}{!}{
    \begin{tabular}{c|c|c|c|c|c|c|c|c|c|c|c|c}
    \toprule
    Settings & \multicolumn{4}{c|}{User Degree Bias} & \multicolumn{4}{c|}{Item Degree Bias} & \multicolumn{4}{c}{\begin{tabular}[c]{@{}c@{}}Item Degree Bias\\ Transductive\end{tabular}} \\
    \midrule
    \multirow{2}[4]{*}{Top-$\mathcal{K}$} & \multicolumn{2}{c|}{NDCG@$1e^{-3}$} & \multicolumn{2}{c|}{Recall@$1e^{-2}$} & \multicolumn{2}{c|}{NDCG@$1e^{-3}$} & \multicolumn{2}{c|}{Recall@$1e^{-2}$} & \multicolumn{2}{c|}{NDCG@$1e^{-3}$} & \multicolumn{2}{c}{Recall@$1e^{-2}$} \\
\cmidrule{2-13}          & top50 & top60 & top50 & top60 & top50 & top60 & top50 & top60 & top50 & top60 & top50 & top60 \\
    \midrule
    NeuMF-Linear                  & 3.47          & 3.94          & 1.32          & 1.60          & 3.19          & 3.67          & 1.24          & 1.51          & 4.10          & 4.44          & 1.46          & 1.68          \\
NeuMF-Linear-L1               & 3.59          & 4.06          & 1.39          & 1.67          & 3.22          & 3.72          & 1.25          & 1.53          & 4.00          & 4.58          & 1.43          & 1.76          \\
NeuMF-Linear-L2               & 3.59          & 4.06          & 1.38          & 1.67          & 3.22          & 3.74          & 1.25          & 1.54          & 4.04          & 4.57          & 1.45          & 1.78          \\
CausPref$^{(--)}$-Linear & 3.48 & 4.01 & 1.31 & 1.61 & 3.96 & 4.18 & 1.40 & 1.64 & 4.28 & 4.75 & 1.59 & 1.82 \\
CausPref$^{(-)}$-Linear          & 3.47          & 3.83          & 1.34          & 1.57          & 4.16          & 4.51          & 1.49          & 1.71          & 4.97          & 5.37          & 1.78          & 2.01          \\
\textbf{CausPref-Linear} & \textbf{4.40} & \textbf{4.86} & \textbf{1.65} & \textbf{1.94} & \textbf{4.84} & \textbf{5.29} & \textbf{1.76} & \textbf{2.04} & \textbf{5.60} & \textbf{6.10} & \textbf{1.99} & \textbf{2.29} \\
NeuMF                & 3.78          & 4.16          & 1.46          & 1.70          & 3.36          & 3.68          & 1.27          & 1.49          & 4.86          & 5.30          & 1.75          & 2.08          \\
NeuMF-L1             & 4.11          & 4.48          & 1.51          & 1.75          & 3.55          & 3.89          & 1.31          & 1.52          & 4.83          & 5.25          & 1.76          & 2.00          \\
NeuMF-L2             & 3.86          & 4.32          & 1.47          & 1.74          & 3.50          & 3.85          & 1.31          & 1.52          & 4.61          & 5.13          & 1.83          & 2.13          \\
NeuMF-dropout & 4.23 & 4.55 & 1.33 & 1.56 & 3.75 & 4.11 & 1.44 & 1.67 & 5.14 & 5.53 & 1.81 & 2.19 \\
IPS            & 3.81 & 4.31 & 1.37 & 1.72 & 3.19 & 3.76 & 1.19 & 1.44 & 4.25 & 4.70 & 1.44 & 1.70 \\
CausE & 3.52 & 4.26 & 1.35 & 1.76 & 3.32 & 3.92 & 1.31 & 1.64 & 4.44 & 4.69 & 1.49 & 1.77 \\
DICE           & 3.71 & 4.06 & 1.52 & 1.73 & 3.75 & 4.13 & 1.36 & 1.62 & 4.78 & 5.26 & 1.70 & 2.03 \\
Light-GCN & 4.63 & 4.90 & 1.40 & 1.56 & 1.26 & 1.43 & 0.48 & 0.59 & 5.32 & 5.64 & 1.80 & 1.99 \\
CausPref$^{(--)}$       & 4.08 & 4.51 & 1.56 & 1.88 & 3.13 & 3.71 & 1.21 & 1.52 & 5.03 & 5.45 & 1.86 & 2.14 \\
CausPref$^{(-)}$       & 2.38          & 2.61          & 0.90          & 1.04          & 2.14          & 2.33          & 0.80          & 0.91          & 3.27          & 3.61          & 1.08          & 1.18          \\
\textbf{CausPref}        & \textbf{4.78} & \textbf{5.21} & \textbf{1.79} & \textbf{2.04} & \textbf{4.40} & \textbf{4.81} & \textbf{1.58} & \textbf{1.82} & \textbf{5.54} & \textbf{6.13} & \textbf{2.04} & \textbf{2.37} \\
    \bottomrule
    \end{tabular}}
  \label{tb1}%
\end{table*}%

\begin{figure}
    \centering \vspace{-10pt}
    \includegraphics[width=0.93\linewidth]{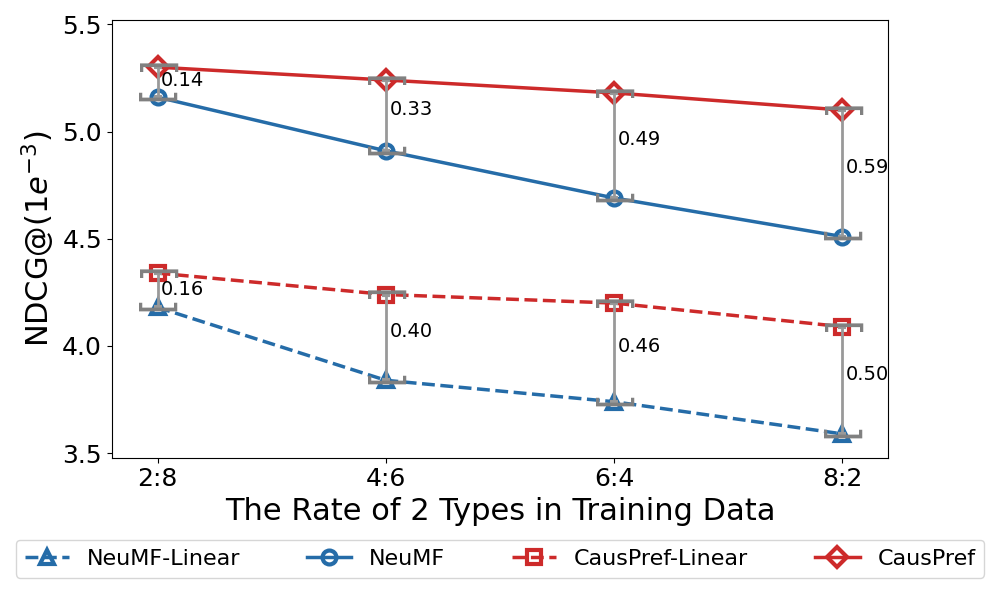}\vspace{-13pt}
    \caption{The  Stability  of  Different Methods  towards  the Change of Distribution Shift for User Feature Bias in the Cloud Theme Click Dataset.}\label{fg3}\vspace{-10pt}
\end{figure}

Here, we consider 5 types of OOD settings that is ubiquitous in real world, as follow,

\begin{itemize}[leftmargin=0.3cm]
    \item User Feature Bias: User groups on online platform usually change over time. For example, the number of female users grows in the International Women's Day.
    \item Item Feature Bias: The item types available on the platform is also time varying. For example, owing to the branding and marketing strategies, the emerging popular product may rapidly occupy a large market share.  
    \item User Degree Bias: The activities of users on platform are often imbalanced. The dominance of behavior data of active users may undesirably affect the recommendation result for inactive users. 
    \item Item Degree Bias: The frequency of item exposure in implicit data follow the long tail distribution, leading to the unfair treatment to unpopular items (less chances to be exposed).
    \item Region Bias: The natural and economic conditions in different regions reveal significant heterogeneity. This lead to distribution shift of users and features when  the recommender system is developed in a new region.
    
\end{itemize}

To simulate the realistic states, 
in Feature Bias settings, 
we first cluster all the samples into two types based on the user or item features, and then adjust the different ratios of data of two types in training/validation and test data; 
In Degree Bias settings that is self-contained in recommendation scenario, 
we randomly sample from the raw dataset for training/validation, 
and form the test data by sampling mainly from the users or items with the lower degree; 
In Region Bias setting, we split the data using the landmark label which is provided in the dataset, i.e. we take data collected in city A for training/validation and city B for test.
In all the settings, we set the proportion of training, validation and test data as $8:1:2$. 
Furthermore, we will focus on the experimental settings where the distribution shift is accompanied by new users/items for generality, that is wide-spread in the real world, unless otherwise specified.

\begin{table*}[tbp]
  \centering
  \caption{The Experimental Results of Region/Feature Bias Settings in the Dinning-Out Dataset.}
  \vspace{-5pt}
  \resizebox{0.85\linewidth}{!}{
    \begin{tabular}{c|c|c|c|c|c|c|c|c|c|c|c|c}
    \toprule
    Settings & \multicolumn{4}{c|}{Region Bias} & \multicolumn{4}{c|}{User Feature Bias} & \multicolumn{4}{c}{Item Feature Bias} \\
    \midrule
    \multirow{2}[4]{*}{Top-$\mathcal{K}$} & \multicolumn{2}{c|}{NDCG@$1e^{-2}$} & \multicolumn{2}{c|}{Recall@$1e^{-2}$} & \multicolumn{2}{c|}{NDCG@$1e^{-2}$} & \multicolumn{2}{c|}{Recall@$1e^{-2}$} & \multicolumn{2}{c|}{NDCG@$1e^{-2}$} & \multicolumn{2}{c}{Recall@$1e^{-2}$} \\
\cmidrule{2-13}          & top20 & top25 & top20 & top25 & top20 & top25 & top20 & top25 & top20 & top25 & top20 & top25 \\
    \midrule
    NeuMF-Linear   & 25.45 & 26.57 & 48.97 & 53.91 & 19.24 & 20.30 & 41.48 & 46.18 & 17.20 & 18.09 & 35.53 & 39.63 \\
    NeuMF-Linear-L1 & 25.44 & 26.57 & 48.97 & 53.91 & 19.25 & 20.30 & 41.50 & 46.18 & 17.48 & 18.35 & 36.28 & 40.26 \\
    NeuMF-Linear-L2 & 25.45 & 26.57 & 48.97 & 53.91 & 19.26 & 20.30 & 41.51 & 46.15 & 17.20 & 18.09 & 35.53 & 39.63 \\
CausPref$^{(--)}$-Linear & 25.37 & 26.58 & 48.90 & 54.01 & 19.41 & 20.39 & 41.89 & 46.32 & 17.49 & 18.48 & 36.05 & 40.54 \\
    CausPref$^{(-)}$-Linear & 24.90 & 25.90 & 48.73 & 53.05 & 19.86 & 20.92 & 42.04 & 46.83 & 17.76 & 18.67 & 36.85 & 40.98 \\
    \textbf{CausPref-Linear} & \textbf{28.39} & \textbf{29.57} & \textbf{54.77} & \textbf{59.95} & \textbf{22.55} & \textbf{23.69} & \textbf{47.61} & \textbf{52.88} & \textbf{22.05} & \textbf{23.20} & \textbf{46.28} & \textbf{51.56} \\
    NeuMF & 26.32 & 27.33 & 51.15 & 55.61 & 19.91 & 20.85 & 42.64 & 46.74 & 18.16 & 19.14 & 37.73 & 42.16 \\
    NeuMF-L1 & 26.34 & 27.35 & 51.21 & 55.66 & 20.23 & 21.13 & 42.39 & 46.57 & 18.28 & 19.25 & 38.00 & 42.48 \\
    NeuMF-L2 & 26.38 & 27.39 & 51.24 & 55.59 & 19.78 & 20.74 & 42.39 & 46.76 & 18.23 & 19.17 & 37.72 & 42.07 \\
    NeuMF-dropout & 26.47 & 27.46 & 51.30 & 55.70 & 20.02 & 20.96 & 42.28 & 46.57 & 18.24 & 19.22 & 37.80 & 42.38 \\

IPS            & 26.05 & 27.21 & 49.81 & 54.56 & 19.52 & 20.54 & 41.67 & 46.33 & 18.77 & 19.59 & 39.12 & 42.87 \\
CausE & 25.47 & 26.62 & 49.08 & 54.17 & 19.24 & 20.37 & 41.44 & 46.40 & 18.78 & 19.69 & 38.32 & 42.45 \\
DICE           & 25.21 & 26.26 & 48.94 & 53.61 & 19.94 & 21.02 & 42.35 & 47.34 & 18.02 & 18.94 & 37.28 & 41.47 \\
 Light-GCN      & 17.14 & 18.52 & 37.83 & 43.96 & 14.14 & 15.39 & 31.69 & 37.36 & 11.78 & 12.81 & 28.75 & 33.46 \\
CausPref$^{(--)}$        & 26.30 & 27.26 & 51.22 & 55.43 & 20.24 & 21.18 & 43.01 & 47.39 & 17.97 & 18.91 & 37.63 & 41.95 \\
    CausPref$^{(-)}$ & 25.58 & 26.62 & 49.96 & 54.51 & 19.89 & 20.83 & 42.38 & 46.68 & 18.38 & 19.30 & 38.60 & 42.81 \\
    \textbf{CausPref} & \textbf{27.27} & \textbf{28.29} & \textbf{53.17} & \textbf{57.82} & \textbf{20.67} & \textbf{21.73} & \textbf{43.50} & \textbf{48.37} & \textbf{19.75} & \textbf{20.77} & \textbf{42.14} & \textbf{46.87} \\
    \bottomrule
    \end{tabular}}%
  \label{tb2}%
\end{table*}%
\subsubsection{Baselines}
For the implementation of proposed CausPref, we take the NeuMF \cite{DBLP:conf/www/HeLZNHC17} as prediction model $\mu$.  
Further, we also replace the MLP layers of NeuMF with a weight matrix to obtain the linear version of the methods, denoted as NeuMF-Linear and CausPref-Linear respectively. 
We choose a variety of baseline models for extensive comparison, including:

\begin{itemize}[leftmargin=0.3cm]
    \item NeuMF/NeuMF-Linear: The vanilla backbone models of CausPref.
    \item CausPref$^{(-)}$/CausPref$^{(-)}$-Linear: The weakened version of CausPref, 
    without the loss $\mathcal{L}_{zero}$ and $\mathcal{L}_{nor}$ that constrain the specific structure properties of user preference. 
    \item CausPref$^{(--)}$/CausPref$^{(--)}$-Linear: Another weakened version of CausPref, 
    that removes the DAG regularizer and only learns the user preference under the reconstruction loss $\mathcal{L}_\mathrm{rec}^N$. 
    \item IPS-based Model: The method that reweighs the BPR loss of training NeuMF according to  inverse propensity score in \cite{yang2018unbiased}.
    \item CausE: The method that trains a NeuMF in a small-scale unbiased data sampled by a well-designed strategy to refine another NeuMF trained in biased  observational data. 
    \item DICE: The method that enhances the robustness despite the distribution shift through learning the disentangled causal embedding based on predefined structure.
    \item Light-GCN: The method exploiting the bonus of GCN model, with the power of ablating the data noise. It is generally regarded as one of the SOTA models.
    \item Regularization Technology: The frequently used approaches including the lasso ($l_1$-norm) \cite{zhao2006model} for sparse feature selection, 
    the $l_2$-regularizer \cite{hoerl1970ridge} that controls the model complexity,
    and the dropout \cite{srivastava2014dropout} amounting to aggregate a group of neural networks. We use them on the prediction model $\mu$.
    
\end{itemize}

For fair comparison, we adapt all the models to utilize the user and item features provided in dataset as input.
And we carry out each experiments for 10 times with different random seeds.

\subsubsection{Metrics} For evaluation, we adopt the $NDCG$ \cite{wang2013theoretical} and $Recall$ \cite{cremonesi2010performance} indices that is widely used in recommendation scenario. 
The top-$\mathcal{K}$ $Recall$ measures the ratio of the item matching with the user in his/her historical record appears in the top-$\mathcal{K}$ of recommended list to that user 
over all test data.
Further, the top-$\mathcal{K}$ $NDCG$ considers the position of the truth item in list and give larger reward for a matching in more forward position.

\subsection{Public Customer-To-Customer Dataset}
Firstly, we study our problem in the public "Cloud Theme Click Dataset"\footnote{The dataset is available from https://tianchi.aliyun.com/dataset/dataDetail?dataId=9716}\cite{du2019sequential} collected in Taobao app of Alibaba representing an important recommendation procedure in the mobile terminal of C2C, including the purchase histories (that we use here) of users for one month before the promotion started.
In this dataset, we evaluate the performance of our CausPref and baselines for User/Item Degree Bias respectively. 
And we consider both the transductive and inductive learning in Item Degree Bias setting. 
The production of experimental data follows  the rules mentioned above.

\begin{figure} 
\centering \vspace{-10pt} 
\includegraphics[width=0.9\linewidth]{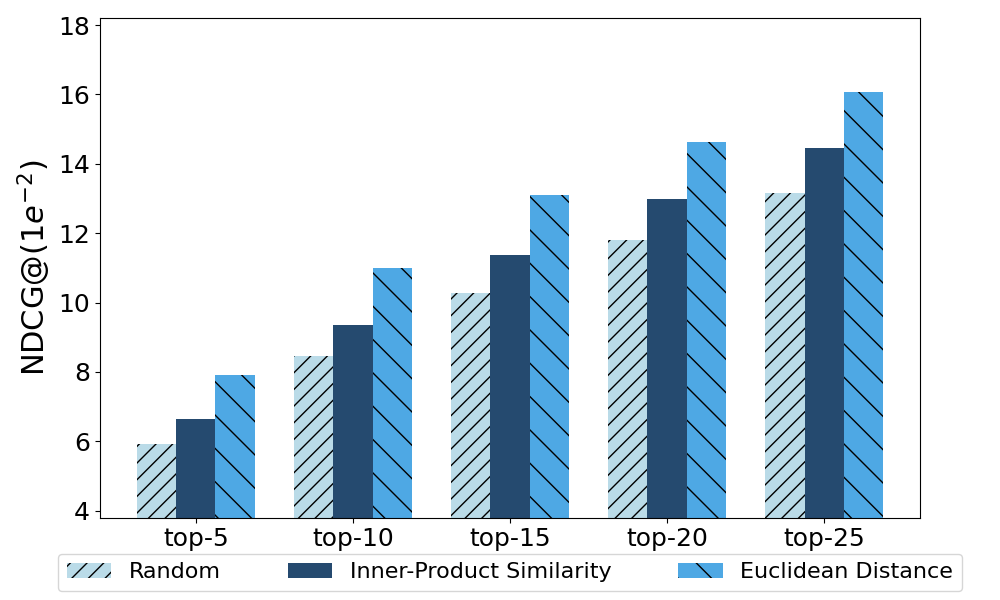}
\centering \vspace{-10pt}
\caption {The Performance of CausPref with Different Negative Sampling Strategies for Item Degree Bias in the Dinning-Out Dataset.}
   \vspace{-10pt}
\label{fg4}
\end{figure} 

From the experimental results reported in Table\ref{tb1}, we find:  
\vspace{-3pt}
\begin{itemize}[leftmargin=0.3cm]
    \item The inductive learning is much more challenging than the transductive learning in the same Bias setting, illustrating that handling new users/items should not be neglected in OOD recommendation.
    \item The neural network can bring the improvement owing to its better capability in modeling nonlinear relationship between variables.
    \item The regularization approaches can help to promote the generalization slightly by alleviating the overfitting issue. In some settings, it even does not work.
    \item The performance of Light-GCN descends severely when the distribution shift happens.
    \item The IPS and CasuE are competitive in some cases, while not in the others, because their de-bias strategy rely on the auxiliary unbiased data that is not always available in practice.
    \item The DICE also performs not steadily, 
    implying its pre-defined causal structure only target the specific bias, not the general distribution shift.
    \item Without the DAG reguralizer, the estimation of user preference easily fits the spurious and unstable patterns in biased data, leading to the failure of CausPref$^{(--)}$/CausPref$^{(--)}$-Linear.
    \item The CausPref$^{(-)}$/CausPref$^{(-)}$-Linear even performs worse than NeuMF/ NeuMF-Linear in some cases, proving the necessity of the specific characteristics of causal structure in recommendation scenario.
    \item The CausPref/CausPref-Linear achieves the best performance with a remarkable improvement for all the metrics in each case, and the gap is larger as the $\mathcal{K}$ increases, demonstrating the superiority of our model.
\end{itemize}

Next, we inspect the stability of algorithm towards the change of distribution shift. In User Feature Bias setting, we fix the ratio of 2 cluster types A:B=2:8 in test data, and adjust it from 2:8 to 8:2 in training data. We report the performances of best model CausPref/CausPref-Linear and basic model NeuMF/NeuMF-Linear in Figure\ref{fg3}. In the I.I.D case (the same ratio for training and test), CausPref/CausPref-Linear is comparable with its backbone, but  has a slight promotion because it can handle the subtle distribution shift easily brought by new users/items. As the distribution shift increases, the advantage of CausPref becomes more significant, verifying that ours can indeed facilitate the stability.

\subsection{Large-scale Business-To-Customer E-Commerce} 
Further, we conduct experiments on a dinning-out dataset collected from a large-scale B2C E-commerce. It provides the raw semantic features of costumers ($100,000+$) and restaurants ($100,000+$), and their connections (the sparsity degree $\leq 1e^{-3}$) if the costumer patronizes the restaurant. We will provide more details about this dataset in appendix. In this dataset, we first evaluate the performance of our CausPref and baselines for User/Item Feature Bias and Region Bias. The ratio of 2 cluster types A:B in training and test data for Feature Bias is 8:2 and 2:8 respectively.

From the experimental results reported in Table\ref{tb2}, we find:
\begin{itemize}[leftmargin=0.3cm]
    \item Similar to the results in Table\ref{tb1}, the CausPref/CausPref-Linear achieves the best performance despite distribution shift in all the cases.
    \item The CausPref-Linear outperforms CausPref. This is because that the linear function is superior in processing the sparse recommendation data\cite{DBLP:conf/www/HeLZNHC17} while the linear model can well capture the underlying relationship between variables.
    \item In Region Bias setting, where the methods is tested almost on the new users/items, 
    our methods still shows the significant superior over baselines.
    We conduct additional experiments to further show our advantage in this case. The results is listed in appendix.
\end{itemize}

Next, we study the impacts of  negative sampling strategies on our CausPref. For the Item Degree Bias, we select the best CausPref-Linear model and try 3 different negative sampling strategies while keeping other components fixed, including random sampling, APS taking the inner-product similarity or Euclidean distance (used in this paper) as distance function respectively. The results in Figure\ref{fg4} then validate the effectiveness of APS algorithm.

\subsection{Interpretable Recommendation Structure}

\begin{figure} [t] 
\centering 
\includegraphics[width=0.9\linewidth]{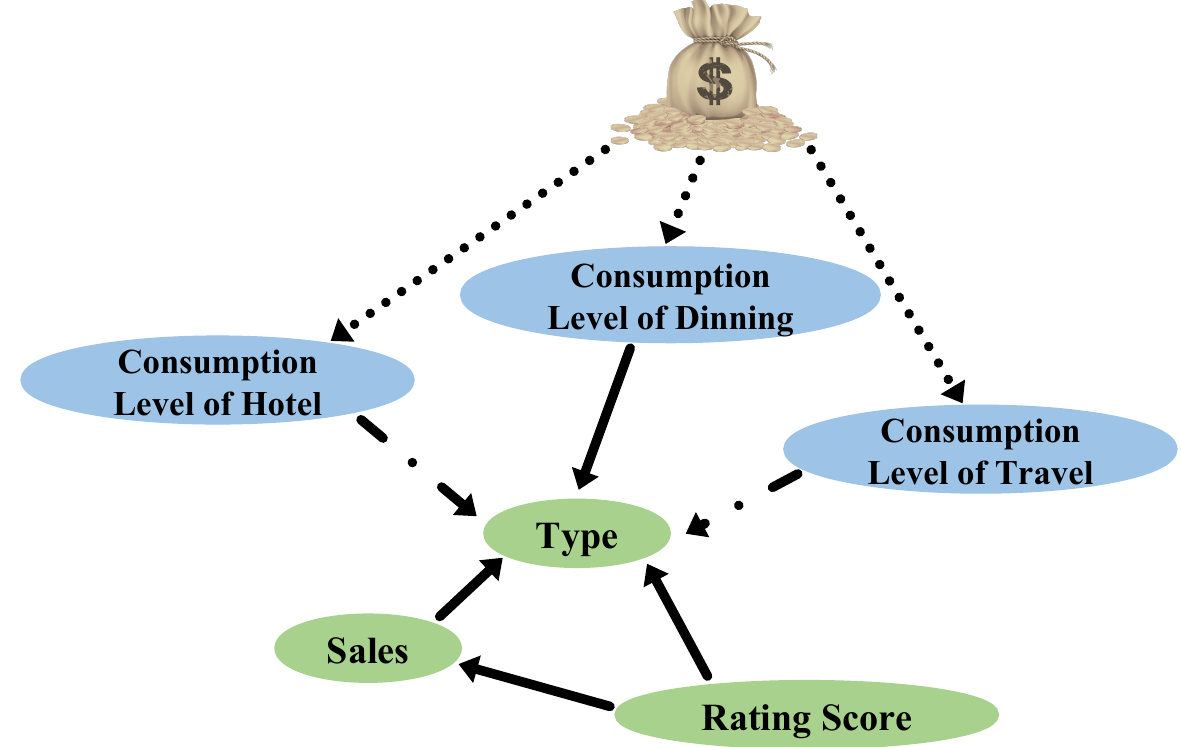} 
\centering
\caption {The Partial Recommendation Structure in Dinning-Out Dataset. The solid/broken/dotted lines are discovered by CausPref, additionally used by traditional methods, and latent unobserved structure respectively.}
\label{fg5}
\vspace{-10pt}
\end{figure}

As been claimed above, our method have improved interpretability if the semantics of features are available. 
Taking partial structure between consumer and restaurant features learnt by CausPref in dinning-out dataset as an example (in Figure\ref{fg5}), 
traditional methods will use the consumer's consumption levels of hotel and travel to predict the type of restaurant, probably 
because of their correlations with the consumption level of dinning-out brought by  unobserved common cause (e.g. economic state). 
However, this spurious correlation may not hold in some other cases. For example, someone who travels a lot may not spend too much on his/her dinning-out. 
According to common sense, we can understand that it is the consumer's expenses on dinning-out, the total sales and rating score of his/her preferred restaurant that determine the type of restaurant he/she would prefer. 

%% file: paragraph/conclusion.tex
\section{Conclusion}
In this work, we study the recommendation problem where the distribution shift exists. 
With the theoretical reflection of OOD generalization for implicit feedback, we summary a paradigm guiding the design of recommender system against the distribution shift.
Following that, we propose a novel causal preference-based recommendation framework called CausPref, that jointly learn the invariant user preference with causal structure from the available positive feedback in observational data and estimates the prediction score with access to the anti-preference negative sampling to catch unobserved signals. The contributions of this paper are well supported by extensive experiments.

%% file: paragraph/appendix.tex
\section*{Appendix}
\subsection{A1. More Comparative Experiment in Inductive Learning with Distribution Shift} 

Limited to space, we report the results in distribution shifted test environment with all the new users/items (i.e. Region Bias Setting) in the main body, which is the most challenging and realistic OOD scenario, being of great value for application. To further illustrate our advantage in this case, we compare results in the test environment with all new instances or the coexistence of new/old instances for the same Item Feature Bias in Dinning-Out Dataset (we keep the other experimental setups be the same). 

We report the test results and relative improvement of CausPref-Linear to the other baselines in Table\ref{tab:addlabel1} and Table\ref{tab:addlabel2} respectively. From them, it is obvious to see that: 1) the increase in the proportion of new users and items makes the OOD problem more difficult, inducing the change of the support of training and test data distribution; 2) ours can relatively bring the more significant benefit for the new user/item, implying our potential in real application. Actually, these phenomenons accord with the comparison of transductive learning and inductive learning in paper.

\subsection{A2. Additional Remarks for Dinning-Out Dataset}

This data collects the users' dinning-out behavior from 2019.01.01 to 2019.12.31.
After sampling, there are total 314,684 users, 206,082 restaurants and 851,325 records. 
Each record represents the costumer (with user\_id `xxx') arrived the restaurant (with poi\_id `yyy') for once consumption. 
This dataset provides the semantic features of user and item, as described in Table\ref{tab3}.
Because the dinning-out behavior indeed happens offline, for a specific user, we select the restaurants in the area around him/her or the area he/she had visited as the item pool for this user in test data.

\begin{table} [b]
\caption{Partial Features in Dinning-Out Dataset.}
\label{tab3}
\begin{tabular}{c|l}
\toprule 
Feature & \multicolumn{1}{c}{Description}\\
\midrule
\multicolumn{2}{c}{User (Costumer)}\\ 
\midrule
user\_id & the ID of user \\
age\_group & the age group of user\\
gender &  male or female \\
married & if he/she is married \\
job & if he/she has a job \\
has\_car & if he/she has a car \\
area\_id\_work & the ID of area where he/she works\\
area\_id\_home & the ID of area where he/she lives \\
mobile\_os & mobile phone system: ios or android\\
consumption\_type & the consumption frequency\\
user\_level\_dinning & the consumption level of dinning-out \\
user\_level\_hotel & the consumption level of hotel \\
user\_level\_travel & the consumption level of travel \\
\midrule
\multicolumn{2}{c}{Restaurant}\\
\midrule
poi\_id& the ID of restaurant\\
category\_id & the type restaurant belongs to \\
business\_area\_id & the ID of area where restaurant locates \\
city\_id & the ID of city where restaurant locates \\
brand\_id & the ID of brand restaurant belongs to \\
ave\_score & the rating score of restaurant \\
avg\_pay & the consumption per person in restaurant \\
rival\_cnt &  the number of restaurants in the same area \\
trade\_cnt &  the total sales volume  \\
\bottomrule
\end{tabular}
\end{table}

\begin{table*}[htbp]
  \centering
  \caption{Testing in The Coexistence of New and Old Uses/Items.}
    \begin{tabular}{c|c|c|c|c}
    \toprule
    Settings & \multicolumn{4}{c}{Item Feature Bias} \\
    \midrule
    \multirow{2}[4]{*}{Top-$\mathcal{K}$} & \multicolumn{2}{c|}{NDCG@$1e^{-2}$} & \multicolumn{2}{c}{Recall@$1e^{-2}$} \\
\cmidrule{2-5}          & top20 & top25 & top20 & top25 \\
    \midrule
    NeuMF-Linear   & 17.20(+28.20\%) & 18.09(+28.25\%) & 35.53(+30.26\%) & 39.63(+30.10\%) \\
    NeuMF-Linear-L1 & 17.48(+26.14\%) & 18.35(+26.43\%) & 36.28(+27.56\%) & 40.26(+28.07\%) \\
    NeuMF-Linear-L2 & 17.20(+28.20\%) & 18.09(+28.25\%) & 35.53(+30.26\%) & 39.63(+30.10\%) \\
    CausPref$^{(--)}$-Linear & 17.49(+26.07\%) & 18.48(+25.54\%) & 36.05(+28.38\%) & 40.54(+27.18\%) \\
    CausPref$^{(-)}$-Linear & 17.76(+24.16\%) & 18.67(+24.26\%) & 36.85(+25.59\%) & 40.98(+25.82\%) \\
    \textbf{CausPref-Linear} & \textbf{22.05} & \textbf{23.2} & \textbf{46.28} & \textbf{51.56} \\
    NeuMF & 18.16(+21.42\%) & 19.14(+21.21\%) & 37.73(+22.66\%) & 42.16(+22.30\%) \\
    NeuMF-L1 & 18.28(+20.62\%) & 19.25(+20.52\%) & 38.00(+21.79\%) & 42.48(+21.37\%) \\
    NeuMF-L2 & 18.23(+20.95\%) & 19.17(+21.02\%) & 37.72(+22.69\%) & 42.07(+22.56\%) \\
    NeuMF-dropout & 18.24(+20.89\%) & 19.22(+20.71\%) & 37.80(+22.43\%) & 42.38(+21.66\%) \\
    IPS   & 18.77(+17.47\%) & 19.59(+18.43\%) & 39.12(+18.30\%) & 42.87(+20.27\%) \\
    CausE & 18.78(+17.41\%) & 19.69(+17.83\%) & 38.32(+20.77\%) & 42.45(+21.46\%) \\
    DICE  & 18.02(+22.36\%) & 18.94(+22.49\%) & 37.28(+24.14\%) & 41.47(+24.33\%) \\
    Light-GCN & 11.78(+87.18\%) & 12.81(+81.11\%) & 28.75(+60.97\%) & 33.46(+54.09\%) \\
    CausPref$^{(--)}$ & 17.97(+22.70\%) & 18.91(+22.69\%) & 37.63(+22.99\%) & 41.95(+22.91\%) \\
    CausPref$^{(-)}$ & 18.38(+19.97\%) & 19.30(+20.21\%) & 38.60(+19.90\%) & 42.81(+20.44\%) \\
    \textbf{CausPref} & \textbf{19.75(+11.65\%)} & \textbf{20.77(+11.70\%)} & \textbf{42.14(+9.82\%)} & \textbf{46.87(+10.01\%)} \\
    \end{tabular}%
  \label{tab:addlabel1}%
\end{table*}%

\begin{table*}[htbp]
  \centering
  \caption{Testing in All New Users/Items.}
    \begin{tabular}{c|c|c|c|c}
    \toprule
    Settings & \multicolumn{4}{c}{Item Feature Bias} \\
    \midrule
    \multirow{2}[4]{*}{Top-$\mathcal{K}$} & \multicolumn{2}{c|}{NDCG@$1e^{-2}$} & \multicolumn{2}{c}{Recall@$1e^{-2}$} \\
\cmidrule{2-5}          & top20 & top25 & top20 & top25 \\
    \midrule
    NeuMF-Linear   & 9.00(+39.33\%) & 9.80(+40.31\%) & 20.97(+48.02\%) & 24.69(+47.79\%) \\
    NeuMF-Linear-L1 & 8.96(+39.96\%) & 9.79(+40.45\%) & 20.86(+48.80\%) & 24.65(+48.03\%) \\
    NeuMF-Linear-L2 & 8.99(+39.49\%) & 9.79(+40.45\%) & 20.96(+48.09\%) & 24.70(+47.73\%) \\
    CausPref$^{(--)}$-Linear & 9.00(+39.33\%) & 9.89(+39.03\%) & 21.30(+45.73\%) & 25.31(+44.17\%) \\
    CausPref$^{(-)}$-Linear & 8.73(+43.64\%) & 9.55(+43.98\%) & 20.24(+53.36\%) & 23.90(+52.68\%) \\
    \textbf{CausPref-Linear} & \textbf{12.54} & \textbf{13.75} & \textbf{31.04} & \textbf{36.49} \\
    NeuMF & 9.13(+37.35\%) & 9.99(+37.64\%) & 21.72(+42.91\%) & 25.56(+42.76\%) \\
    NeuMF-L1 & 9.61(+30.49\%) & 10.49(+31.08\%) & 22.67(+36.92\%) & 26.72(+36.56\%) \\
    NeuMF-L2 & 9.41(+33.26\%) & 10.35(+32.85\%) & 22.13(+40.26\%) & 26.39(+38.27\%) \\
    NeuMF-dropout & 9.21(+36.16\%) & 10.07(+36.54\%) & 22.05(+40.77\%) & 25.95(+40.62\%) \\
    IPS   & 10.33(+21.39\%) & 11.24(+22.33\%) & 24.51(+26.64\%) & 28.71(+27.10\%) \\
    CausE & 9.78(+28.22\%) & 10.61(+29.59\%) & 23.17(+33.97\%) & 27.23(+34.01\%) \\
    DICE  & 7.88(+59.14\%) & 8.70(+58.05\%) & 19.35(+60.41\%) & 23.41(+55.87\%) \\
    Light-GCN & 3.20(+291.88\%) & 4.22(+225.83\%) & 11.56(+168.51\%) & 16.18(+125.53\%) \\
    CausPref$^{(--)}$ & 9.15(+37.05\%) & 9.98(+37.78\%) & 21.86(+41.99\%) & 25.70(+41.98\%) \\
    CausPref$^{(-)}$ & 7.85(+59.75\%) & 8.59(+60.07\%) & 18.11(+71.40\%) & 21.50(+69.72\%) \\
    \textbf{CausPref} & \textbf{10.56(+18.75\%)} & \textbf{11.56(+18.94\%)} & \textbf{26.03(+19.25\%)} & \textbf{30.75(+18.67\%)} \\
    \end{tabular}%
  \label{tab:addlabel2}%
\end{table*}%